\pdfoutput=1

\documentclass[11pt]{article}

\usepackage[]{acl}

\usepackage{times}
\usepackage{latexsym}
\usepackage{xspace}
\usepackage{booktabs}
\usepackage{tabularx}
\usepackage{multicol}

\usepackage[T1]{fontenc}

\usepackage[utf8]{inputenc}
\usepackage{inconsolata}

\usepackage{microtype}
\usepackage{hyperref}

\usepackage{graphicx}
\usepackage[normalem]{ulem}

\newcommand{\packageName}{\mbox{\textsc{Riveter}}\xspace} 
\newcommand{\emoji}{\includegraphics[width=1em,trim=0 4cm 0 0]{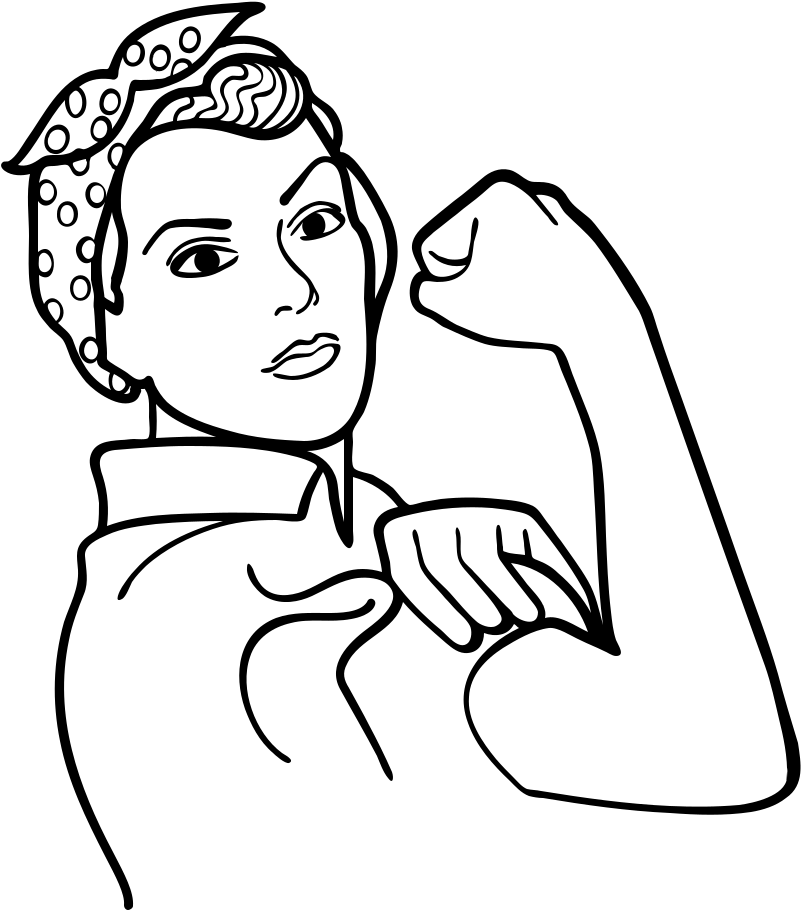}}
\newcommand{\emojiLarge}{\includegraphics[width=1.5em,trim=0 5cm 0 0]{rosieTheRiveter2.png}\vspace{0.25em}}

\newcommand{\packageNameWithEmoji}{\textsc{Riveter}\hspace{.15em}\emoji\xspace}

\title{\packageName\emojiLarge\\Measuring Power and Social Dynamics Between Entities}

\newcommand{\aspace}{\hspace{2em}}
\newcommand{\cmu}{$^\heartsuit$}
\newcommand{\aitwo}{$^\clubsuit$}
\newcommand{\uw}{$^\diamondsuit$}
\newcommand{\emory}{$^\spadesuit$}
\newcommand{\jhu}{$^\dagger$}

\author{Maria Antoniak\aitwo \aspace Anjalie Field\jhu \aspace Jimin Mun\cmu \aspace Melanie Walsh\uw \\
\textbf{Lauren F. Klein\emory \aspace Maarten Sap\cmu\aitwo}
\vspace{.3em}\\
\small{\aitwo Allen Institute for AI \aspace \jhu Johns Hopkins University \aspace \cmu Carnegie Mellon University}\\
\small{\uw University of Washington \aspace \emory Emory University}\vspace{.3em}\\
\url{http://github.com/maartensap/riveter-nlp}
 }

\begin{document}

\maketitle

\begin{abstract}
\packageName provides a complete easy-to-use pipeline for analyzing verb connotations associated with entities in text corpora. 
We prepopulate the package with connotation frames of sentiment, power, and agency, which have demonstrated usefulness for capturing social phenomena, such as gender bias, in a broad range of corpora.
For decades, lexical frameworks have been foundational tools in computational social science, digital humanities, and natural language processing, facilitating multifaceted analysis of text corpora.
But working with verb-centric lexica specifically requires natural language processing skills, reducing their accessibility to other researchers.
By organizing the language processing pipeline, providing complete lexicon scores and visualizations for all entities in a corpus, and providing functionality for users to target specific research questions, \packageName greatly improves the accessibility of verb lexica and can facilitate a broad range of future research.

\end{abstract}

\section{Introduction}

\begin{figure}[t]
    \centering
    \includegraphics[width=\columnwidth]{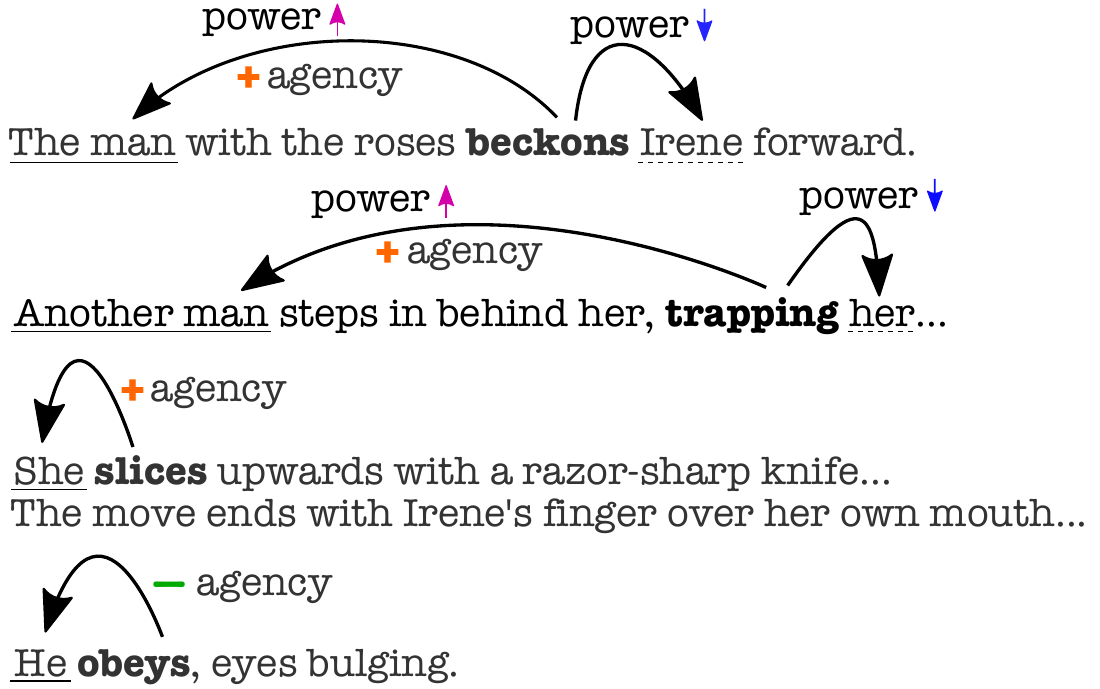}
    \caption{Figure from \citet{sap2017connotation} illustrating power and agency connotation frames extracted on an excerpt from the \textit{Sherlock Holmes} (2009) film screenplay. Each connotation frame pertains to a \textbf{verb predicate} and its \underline{agent} and \dashuline{theme}.}
    \label{fig:introFig}
\end{figure}

Language is a powerful medium that intricately encodes
social dynamics between people, such as perspectives, biases, and power differentials \citep{Fiske1993controlling}.
When writing, authors choose how to \textit{portray} or \textit{frame} each person in a text, highlighting certain features \citep{entman1993framing} to form larger arguments \citep{Fairhurst2005ReframingTA}.
For example, in the screenplay for the 2009 film \textit{Sherlock Holmes}, the authors dramatize a sudden reversal of power by playing on gender stereotypes.
First, they describe how  ``the man with the roses \textbf{beckons} Irene forward'' (Figure \ref{fig:introFig}), which portrays the character Irene Adler as being lured by the man. 
After she is trapped, ``she \textbf{slices} upward with a razor-sharp knife,'' reversing the power dynamic. 
Here, specific word choices shape and then challenge the viewers' expectations about how the interaction is presumed to unfold. 

More broadly, an author's word choices can not only communicate important details about a character or narrative but can also reveal larger social attitudes and biases \cite{Blackstone2003-vb,Cikara2009-rk}, shape readers' opinions and beliefs about social groups \cite{behm2008mean}, as well as act as powerful mechanisms to persuade or to induce empathy \cite{Smith1996-ac,Keller2003-wa}.
Studying these word choices across large datasets can illuminate domain-specific patterns of interest to scholars in the humanities and social sciences.

Examining verbs---\textit{who} does \textit{what} to \textit{whom}?---is one established approach for measuring the textual portrayal of people and groups of people. 
Each verb carries connotations that can indicate the social dynamics at play between the subject and object of the verb.
Connotation frames \citep{rashkin-etal-2016-connotation,sap2017connotation} capture these dynamics by coding verbs with directional scores.
For example, these frames might label verbs with who holds power and who lacks power, or who is portrayed with positive or negative sentiment.
In the \textit{Sherlock Holmes} scene description, ``Another man steps in behind her, \textbf{trapping her},'' the verb \textit{to trap} implies that ``the man'' has more power over ``her'' \cite[Figure \ref{fig:introFig};][]{sap2017connotation}. 
These verb lexica have been used successfully to study portrayals in many diverse contexts including films \citep{sap2017connotation}, online forums \citep{antoniak2019narrative}, text books \citep{lucy2020content}, Wikipedia \citep{Park_Yan_Field_Tsvetkov_2021}, and news articles \citep{field2019contextual}.

Lexica in general are extremely popular among social science and digital humanities scholars. They are interpretable and intuitive \citep{grimmer2013text}, especially when compared with black-box classification models, and continue to be a go-to resource \cite[e.g., LIWC;][]{liwc2015}.
However, verb-based lexica pose specific technical hurdles for those less experienced in software engineering and natural language processing (NLP).
These lexica require core NLP skills such as traversing parse trees, identifying named entities and references, and lemmatizing verbs to identify matches. 
At the same time, the larger research questions motivating their usage require deep domain expertise from the social sciences and humanities.

To meet this need, we introduce \textbf{\packageNameWithEmoji},\footnote{The name Riveter is inspired by ``Rosie the Riveter,'' the allegorical figure who came to represent American women working in factories and at other industrial jobs during World War II. Rosie the Riveter has become an iconic symbol of power and shifting gender roles---subjects that the Riveter package aims to help users measure and explore by combining (or \textit{riveting}) components into a pipeline.} which includes tools to use, evaluate, and visualize verb lexica, enabling researchers to measure power and other social dynamics between entities in text.
This package includes a pipeline system for importing a lexicon, parsing a dataset, identifying people or entities, resolving coreferences, and measuring patterns across those entities.
It also includes evaluation and visualization methods to promote grounded analyses within a targeted dataset.
We release \packageName as a Python package, along with Jupyter notebook demonstrations and extensive documentation aimed at social science and humanities researchers.

To showcase the usefulness of this package, we describe two case studies: (1) power differentials and biases in GPT-3 generated stories and (2) gender-based patterns in the novel \textit{Pride and Prejudice}.
The first study provides a proof-of-concept; dyads with predetermined power differentials are used to generate stories, and we are able to detect these distribution shifts using \packageName.
The second study zooms in on a particular author, text, and social setting, examining how a 19th century novelist both portrayed and subverted gender roles.
These case studies highlight the diverse contexts and research questions for which this package can be used across human and machine-generated text, and across the social sciences and the humanities.

\begin{table*}[t]
    \centering
    \small
    \begin{tabular}{@{}lp{10cm}@{}}
    \toprule
        Work & Usage \\ \midrule
        \citet{rashkin-etal-2016-connotation} & Analyzing political leaning and bias in news articles. \\  
        \citet{sap2017connotation} & Analyzing gender bias in portrayal of characters in movie scripts. \\
        \citet{rashkin-etal-2017-multilingual} & Analyzing public sentiment (and multilingual extension of \citet{rashkin-etal-2016-connotation}) \\ 
        \citet{volkova2018misleading} & Improving the detection of fake news \& propaganda.\\
        \citet{ding2018weakly} & Detecting affective events in personal stories. \\ 
        \citet{field2019contextual} & Analyzing power dynamics of news portrayals in \#MeToo stories. \\ 
        \citet{antoniak2019narrative} & Analyzing the power dynamics in birthing stories online. \\
        \citet{lucy2020content} & Analyzing the portrayal of minority groups in textbooks.\\
        \citet{mendelsohn2020framework} & Analyzing the portrayal of LGBTQ people in the New York Times.\\
        \citet{ma2020powerTransformer} & Text rewriting for mitigating agency gender bias in movies.\\
        \citet{Park_Yan_Field_Tsvetkov_2021} & Comparing affect in multilingual Wikipedia pages about LGBT people \\ 
        \citet{lucy-bamman-2021-gender} & Analyzing gender biases in GPT3-generated stories.\\
        \citet{gong2022gender} & Quantifying gender biases and power differentials in Japanese light novels \\
        \citet{saxena2022unpacking} & Examining latent power structures in child welfare case notes \\
        \citet{borchers2022looking} & Measuring biases in job advertisements and mitigating them with GPT-3  \\
        \citet{stahl2022prefer} & Joint power-and-agency rewriting to debias sentences. \\
        \citet{wiegand2022identifying} & Identifying implied prejudice and social biases about minority groups  \\
        \citet{giorgi2023author} & Examining the portrayal of narrators in moral and social dilemmas\\

         \bottomrule
    \end{tabular}
    \caption{Examples of usages of connotation frames in NLP and CSS literature.}
    \label{tab:usage-examples}
\end{table*}


\section{Background: Verb Lexica \& Connotation Frames}
\newcommand{\panini}{P\= a\d nini\xspace}
\newcommand{\event}{\textsc{Event}\xspace}
\newcommand{\agent}{\textsc{Agent}\xspace}
\newcommand{\theme}{\textsc{Theme}\xspace}

\begin{figure}[t]
    \centering
    \includegraphics[width=\columnwidth]{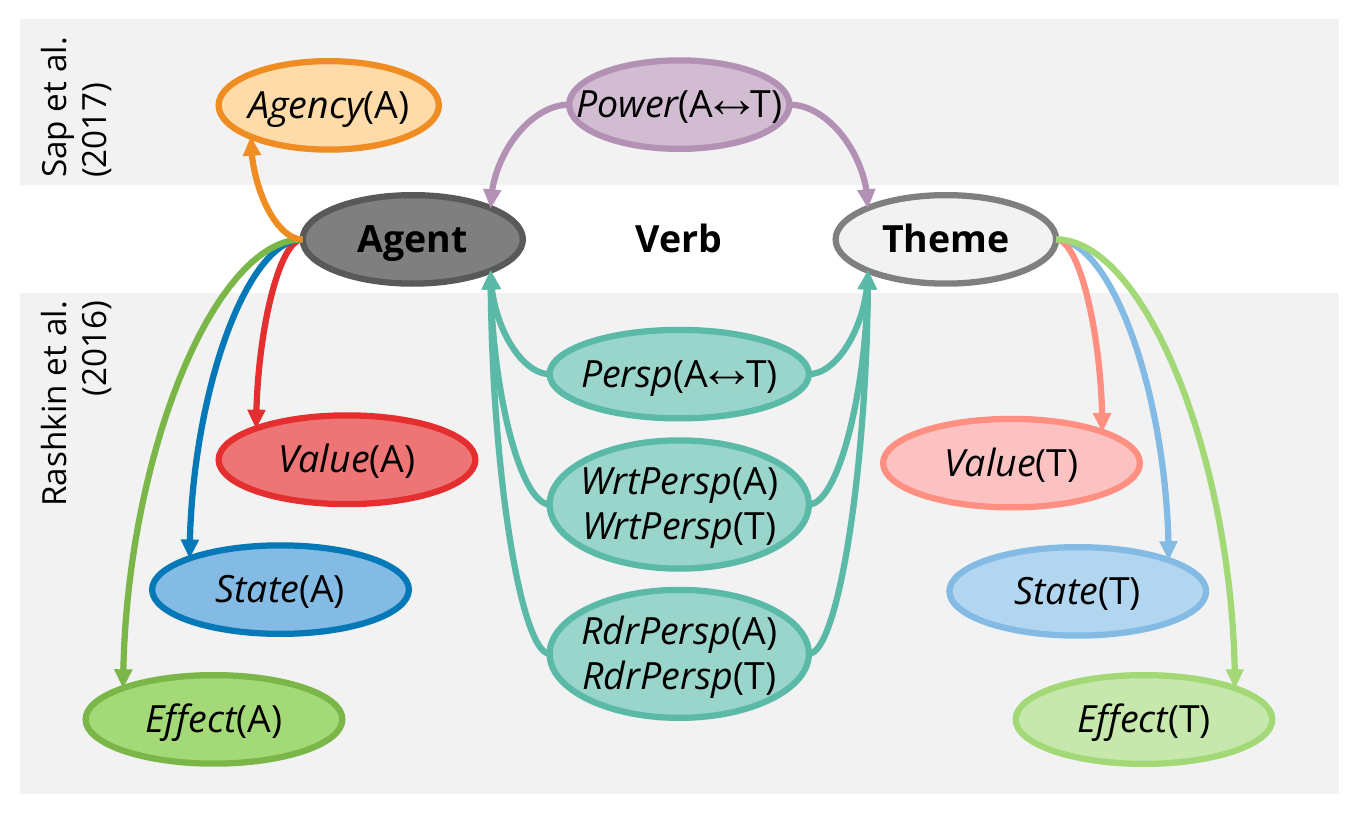}
    \caption{A verb predicate can connote various sentiment, power, and agency levels for its agent and theme; connotation frames distill these into six relation types (\S\ref{ssec:dimenion-definitions}; each type is colored in a different hue).
    }
    \label{fig:frameExample}
\end{figure}

\subsection{Verb Predicates \& Frame Semantics}
Understanding the events and states described in sentences, i.e., \textit{who} does \textit{what} to \textit{whom}, has been a central question of linguistics since first conceptualized by Indian grammarian \panini between the 4th and 6th century BCE.
Today, verb predicates and their relation to other words in a sentence are still a key focus of linguistic analyses, e.g., dependency parsing \cite{tesniere2015elements,nivre2010dependency} and semantic role labeling \cite{gildea2002automatic}.

To model how one understands and interprets the events in a sentence, \citet{fillmore1976frame} introduced \textit{frame semantics}, arguing that understanding a sentence involves knowledge that is evoked by the concepts in the sentence.
This theory inspired the task of \textit{semantic role labeling} \cite{gildea2002automatic}, which categorizes how words in a sentence relate to the main verb predicate 
via frame semantics.
This task defines \textit{thematic roles} with respect to an \event (i.e., the verb predicate): the \agent that causes the \event (loosely, the subject of the verb), and the \theme that is most directly affected by the \event (loosely, the object of the verb).

\subsection{Connotation Frames of Sentiment, Power, and Agency}
While frame semantics was originally meant to capture broad meaning that arises from interpreting words in the context of what is known \cite{fillmore1976frame}, many linguistic theories have focused solely on denotational meaning \cite{baker1998berkeley,palmer2005proposition}, i.e., examining only what is present in the sentence.
In contrast, the \textit{implied or connoted meaning} has received less attention, despite being crucial to interpreting sentences.

\textit{Connotation frames}, introduced by \citet{rashkin-etal-2016-connotation}, were the first to model the connotations of verb predicates with respect to an \agent and \theme's value, sentiment, and effects (henceforth, sentiment connotation frames).
Shortly thereafter, \citet{sap2017connotation} introduced the \textit{power and agency connotation frames} (Figure \ref{fig:frameExample}), which model the power differential between the \agent and the \theme, as well as the general agency that is attributed to the \agent of the verb predicate.\footnote{While the value, sentiment, effects, and power relations require a verb to be transitive, the agency dimension is present with intransitive verbs as well.}

For both sets of connotation frames, the authors released a lexicon of verbs with their scores.
Verbs were selected based on their high usage in corpora of choice: frequently occurring verbs from a corpus of New York Times articles \cite{sandhaus2008new} for sentiment connotation frames, and frequently occurring verbs in a movie script corpus \cite{gorinski2015movie} for the power and agency frames.
Each verb was annotated for each dimension by crowdworkers with \agent and \theme placeholders (``\textit{X implored Y}'').

Since their release, connotation frames have been of increasing interest to researchers working in disciplines like cultural analytics and digital humanities communities.
They have given these researchers a flexible and interpretable way to examine the framing of interpersonal dynamics across a wide range of datasets and research questions (Table \ref{tab:usage-examples}).
Additionally, the frames have been incorporated into the 2023 edition of the textbook \textit{Speech and Language Processing} \cite{Jurafsky2023slp}.

\makeatletter
\newcommand{\negative}{\mathbin{\mathpalette\make@circled-}}
\newcommand{\equal}{\mathbin{\mathpalette\make@circled=}}
\newcommand{\positive}{\mathbin{\mathpalette\make@circled+}}
\newcommand{\make@circled}[2]{%
  \ooalign{$\m@th#1\smallbigcirc{#1}$\cr\hidewidth$\m@th#1#2$\hidewidth\cr}%
}
\newcommand{\smallbigcirc}[1]{%
  \vcenter{\hbox{$\m@th#1\bigcirc$}}%
}
\makeatother

\subsection{Connotation Frame Dimensions}\label{ssec:dimenion-definitions}
Given a predicate verb $v$ describing an \event and its \agent $a$ and \theme $t$, connotation frames capture several implied relations along sentiment, power, and agency dimensions. 
Each of these relations has either a positive $\positive$, neutral $\equal$, or negative $\negative$ polarity.
We describe here a set of six example relations included in \packageName; for a fuller discussion of each of these relations and their definitions, see \citet{rashkin-etal-2016-connotation} and \citet{sap2017connotation}.

\paragraph{\textit{\textcolor[HTML]{7AB648}{Effect}}} denotes whether the event described by $v$ has a positive or negative effect on the agent $a$ or the theme $t$. For example, in Figure \ref{fig:introFig}, another man ``trapping'' Irene has a negative effect on her (\textit{Effect}$(t)=\negative$).

\paragraph{\textcolor[HTML]{E52E2F}{\textit{Value}}} indicates whether the agent or theme are presupposed to be of value by the predicate $v$. For example, when someone ``guards'' an object, this presupposes that the object has high value (\textit{Value}$(t)=\positive$).

\paragraph{\textcolor[HTML]{0578B7}{\textit{State}}} captures whether the likely mental state of the \agent or \theme as a result of the \event. For example, someone ``suffering'' likely indicates a negative mental state (\textit{State$(a)=\negative$}).

\paragraph{\textcolor[HTML]{5ABAA7}{\textit{Perspective}}} is a set of relations that describe the sentiment of the \agent towards the \theme and vice versa (\textit{Persp}$(a\leftrightarrow t)$). 
It also describes how the writer perceives the \agent and \theme (\textit{WrtPersp}$(a)$, \textit{WrtPersp$(t)$}), as well as how the reader likely feels towards them (\textit{RdrPersp}$(a)$, \textit{RdrPersp}$(t)$).

\paragraph{\textcolor[HTML]{B391B5}{\textit{Power}}} distills the power differential between the \agent and \theme of the \event (denoted as \textit{Power}$(a\leftrightarrow t)$ for shorthand). For example, when a man ``traps'' Irene, he has power over Irene (\mbox{\textit{Power}$(a)=\positive$} and \mbox{\textit{Power}$(t)=\negative$}). In the implementation of \packageName, we convert the positive $\positive$, neutral $\equal$, or negative $\negative$ polarities into categorical scores ($\{-1, 0, +1\}$), as described in \S\ref{subsection:lexicon-loading}, to facilitate aggregation over entities.

\paragraph{\textcolor[HTML]{EF8D22}{\textit{Agency}}} denotes whether the \agent of the \event has agency, i.e., is decisive and can act upon their environment. 
For example, Irene ``slicing'' connotes high agency (\mbox{\textit{Agency}$(a)=\positive$}), whereas the man ``obeying'' connotes low agency (\textit{Agency}$(a)=\negative$).
We convert these categories to numbers as described for \textit{Power} above.

\begin{figure*}[t]
    \centering
    \includegraphics[width=\textwidth]{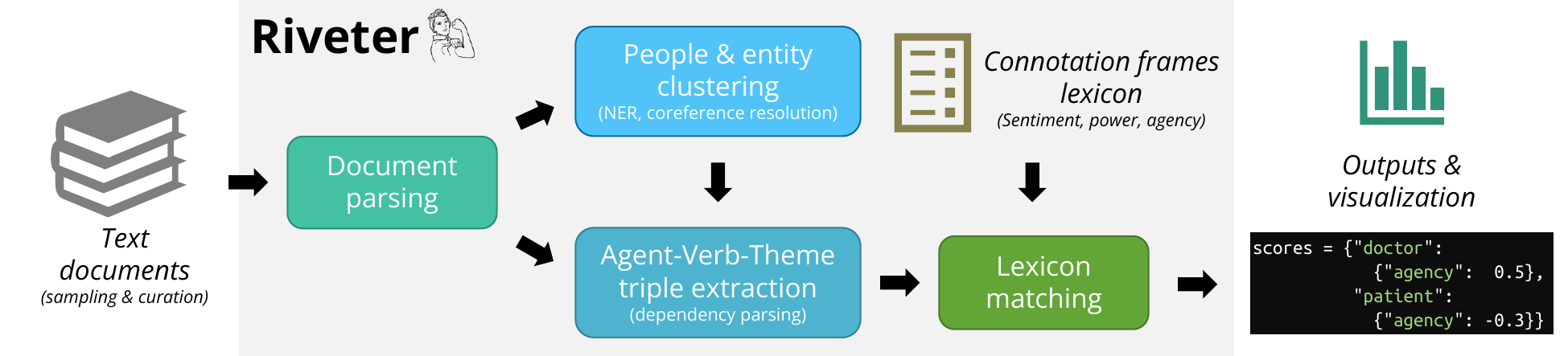}
    \caption{A visualization of the \packageName pipeline components and their connections.}
    \label{fig:systemDiagram}
\end{figure*}

\section{\packageNameWithEmoji Design \& Implementation}

\subsection{Challenges Addressed}

Unlike lexica that require only string matching, verb lexica indicating relations between the \agent and \theme also require parsing, lemmatization, named entity recognition, and coreference resolution.
These are standard pieces of NLP pipelines, but each piece requires background knowledge in linguistics, NLP, and algorithms that inform library choices and merging of outputs; this can pose a challenge for researchers without extensive NLP knowledge or training.
\packageName substantially lowers the implementation burden and text-processing knowledge required for using verb lexica by addressing the following three challenges.

\paragraph{Familiarity with Using NLP Tools}
The increasing availability of NLP packages has resulted in numerous existing packages for core NLP pipelines.
We reviewed the performance (considering accuracy, speed, and ease of installation) of available tools and pre-selected optimal text processing pipelines for \packageName, eliminating the need for users to be familiar with and decide between available text processing tools.
We also provide documentation on incorporated packages and extensive demonstrations.

\paragraph{Interfacing between NLP Tools}

Even if one is familiar with individual tools, like parsers or entity recognizers, connecting outputs from one tool to another tool requires an additional engineering skill set.
Traversing a parse tree to find semantic triples and then matching these triples to clusters from a coreference resolution engine is not a straightforward process for a researcher with less expertise in programming and software engineering.
To address this challenge, we (a) provide a system that connects these pipeline pieces for the user while also (b) providing functionality to explore the outputs of each individual system.

\paragraph{Interpreting Results}

Lexical methods can offer flexibility and interpretability not found in other NLP methods, but even so, validating and exploring lexical results can be challenging.
Proper validation is not consistent even in NLP research using lexicon-based methods \citep{antoniak-mimno-2021-bad}.
To address this challenge, we provide methods to explore the results numerically and visually, enabling users to quickly produce plots, calculate aggregate scores, identify contributing verbs and documents, and measure their results' stability.

\subsection{System Description}

Illustrated in Figure~\ref{fig:systemDiagram}, \packageName takes in a set of documents as input, and returns a set of scores for each entity appearing as an \agent or \theme in the target dataset.
Under the hood, \packageName parses the documents, resolves coreference clusters, finds entity mentions, extracts \agent-\event-\theme triples, and computes lexicon scores. 
We verify our implementation through hand-constructed unit tests and testing of large and small corpora.
We describe each of these components below.

\paragraph{Named Entity Recognition and Coreference Resolution}
Our package first parses a given document to find clusters of mentions that relate to entities.
We extract general coreference clusters, which we cross-reference with mentions of entities labeled by a named entity recognition (NER) system, as well as a list of general people referents (containing pronouns, professions, etc).
Coreference cluster mentions that overlap with a mention of an entity are then passed to the verb and relation identification module.

In our implementation, we currently use spaCy for NER, and the spaCy add-on \textit{neuralcoref}\footnote{\url{https://github.com/huggingface/neuralcoref}} library for coreference resolution.
These libraries are well-supported, have fast run-times relative to similar systems, and do not require GPU access, which are not accessible to many researchers, especially outside of computing disciplines.

\paragraph{Lexicon Loading}
\label{subsection:lexicon-loading}
We include two lexica by default: connotation frames from \citet{rashkin-etal-2016-connotation} and frames of power and agency from \citet{sap2017connotation}.
These lexica come included in the package, and the user can select between the lexica and their dimensions (see \S\ref{ssec:dimenion-definitions} for dimension descriptions), as well as use their own custom lexica.

For the verb lexicon from \citet{rashkin-etal-2016-connotation} and for custom lexicons, each verb has a numerical score (ranging from $-1$ to $1$) for each of \agent and \theme.
If a lexicon only has scores for \agent or \theme, the other scores are set to $0$.
For the verb lexicon from \citet{sap2017connotation}, we convert the categorical labels to numerical scores, so that each verb has a score of $+1$, $0$, or $-1$ for each of the \agent and \theme.

For example, in the verb lexicon from \citet{sap2017connotation}, the verb ``amuse'' is labeled as having positive agency (\mbox{\textit{Agency}$(a)=\positive$}) and power for the \theme (\mbox{\textit{Power}$(a)=\negative$}, \mbox{\textit{Power}$(t)=\positive$}).
In this lexicon, agency was coded only for the \agent, so we convert the categorical label to a score of $+1$ for the \agent and $0$ for the \theme.
For power, we convert the categorical label to $-1$ for the \agent and $+1$ for the \theme.

\packageName also allows the use of custom lexica.
We include a loading function for any verb lexicon formatted in the style of \citet{rashkin-etal-2016-connotation}.
This requires a file listing verbs and their agent and theme scores, which should be positive and negative numbers.
This functionality is especially important when using previous lexica on new datasets, as this allows users to customize those lexica for new contexts, simply by updating or adding to the included lexicon files.

\paragraph{Verb Identification and Entity Relation}
We extract the lemmas of all verbs and match these to the lexicon verbs.
After identifying semantic triples (the \agent or subject (\textit{nsubj}) and \theme or direct object (\textit{dobj}) of each verb) using the spaCy dependency parser, we search for matches to the NER spans identified in \S\ref{subsection:lexicon-loading}.
We track the frequencies of these for the canonical entity, using the converted scores.

\paragraph{Exploration and Visualization}
We provide functionality for users to easily view lexicon scores for entities in their input text. To maximize utility, we focus on facilitating analyses established in prior work (e.g., Table~\ref{tab:usage-examples}). This functionality includes:

\begin{itemize}
    \item retrieving the overall verb lexicon scores for all entities identified in the entire input corpora (\texttt{get\_score\_totals}),
    \item retrieving the overall verb lexicon scores for all entities identified in a specific document (\texttt{get\_scores\_for\_doc}),
    \item generating bar plots of scores for entities in the dataset (e.g., filtering for the top-scored entities) (\texttt{plot\_scores}) or in a specific document (\texttt{plot\_scores\_for\_doc}).
\end{itemize}

We additionally provide functionality to reduce the opacity of lexicon scores and allow users to examine specific findings in more depth or conduct error checking. These functions include:

\begin{itemize}    
    \item generating heat map plots for the verbs most frequently used with a user-specified entity (\texttt{plot\_verbs\_for\_persona}),
    \item retrieving all mentions associated with a specified entity, after co-reference resolution (\texttt{get\_persona\_cluster}),
    \item retrieving various additional counts, including number of lexicon verbs in a document, all entity-verb pairs in a document, number of identified entities, etc.
    \item retrieving all documents that matched an entity or verb (\texttt{get\_documents\_for\_verb}),
    \item bootstrapping the dataset to examine stability across samples (\texttt{plot\_verbs\_for\_persona}). 
\end{itemize}

\section{Case Studies Across Cultural Settings}

\subsection{Machine Stories with Power Dyads}
\label{subsec:power-dyads}
As our first case study, we examine lexicon-based power differentials in machine-generated stories about two characters with a predetermined power asymmetry, as in [``doctor'', ``patient''], [``teacher'', ``student''], and [``boss'', ``employee''].
By generating stories about entities with predetermined power asymmetries, this serves as a proof-of-concept for \packageName; we expect to measure power scores in the predetermined directions.

\begin{figure}[t]
    \centering
    \includegraphics[width=0.9\columnwidth]{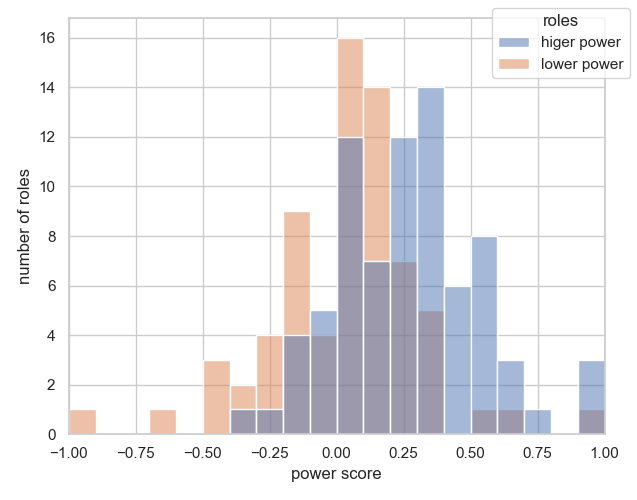}
    \caption{Number of roles with corresponding power scores color coded by assignment of higher and lower power roles over 85 short stories sampled from GPT-3.5.}
    \label{fig:power_score_plot}
\end{figure}

\begin{figure}[t]
    \centering
    \includegraphics[width=0.9\columnwidth]{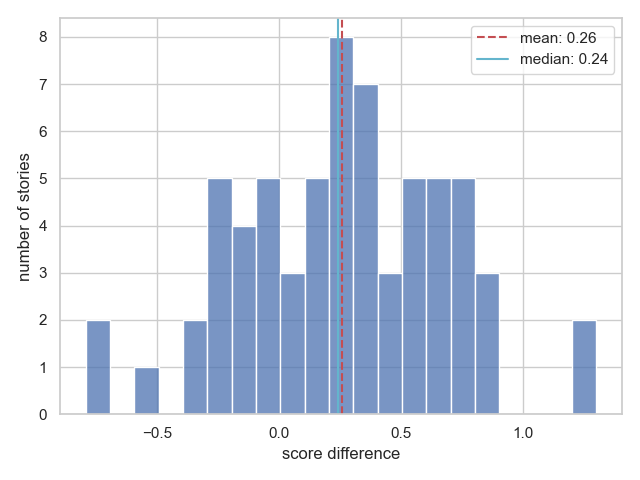}
    \caption{Number of stories with corresponding power score differences calculated by taking 
    \mbox{\textit{Power}$(e^+) - $\textit{Power}$(e^-)$} in each generated story where $e^+$ and $e^-$ are pairs of entities predetermined to have high and low power roles (e.g., [``doctor''$-$``patient'']). }
    \label{fig:score_diff_plot}
\end{figure}

Given a set of 32 pairs of roles, we obtain 85 short stories from GPT-3.5 \cite{Ouyang2022InstructGPT}
(see Appendix \ref{sec:appendix-gpt3} for details).
We then scored each of the characters with assigned roles using \packageName and aggregated the scores for higher power roles and lower power roles using their names given by GPT-3.5. 

\paragraph{Results}
As seen in Figure~\ref{fig:power_score_plot}, higher power roles have a distribution shifted toward greater power scores than lower power roles. 
The mean score of the higher power roles was $0.265$ and that of lower power roles was $0.0364$. 
A t-test also shows statistical significance ($p<0.05$).
From these results we can conclude that the stories generated by GPT-3.5 reflect the power dynamics in the relations given in the prompt, and our framework captures this expected phenomena. 

The differences in power scores show similar results in Figure~\ref{fig:score_diff_plot}. 
These differences were calculated only for the stories where both higher and lower power figures had been scored. 
The mean and median were both positive, $0.26$ and $0.24$.

Analyzing the stories with both positive and negative score differences (see Table~\ref{tab:gpt3_stories} in the Appendix) further confirms the results of our framework. 
For example, the third story of the table shows negative score difference between higher powered agent and lower powered agent. 
However, looking at it more closely we can see that despite the assigned role as an interviewee, Emily shows greater power.
In the sentence ``Paul thanked Emily for her time and wished her luck,'' both \textit{thank} and \textit{wish} from our lexicon give Emily greater power than Paul. 
Thus we can analyze the power dynamics between characters more accurately, taking into account more context than just assigned roles.

\begin{figure}[t]
    \centering
    \includegraphics[width=7cm]{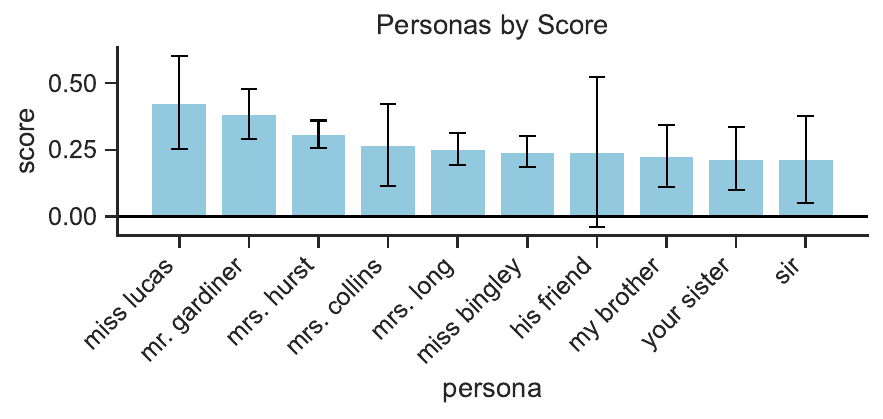}
    \includegraphics[width=7cm]{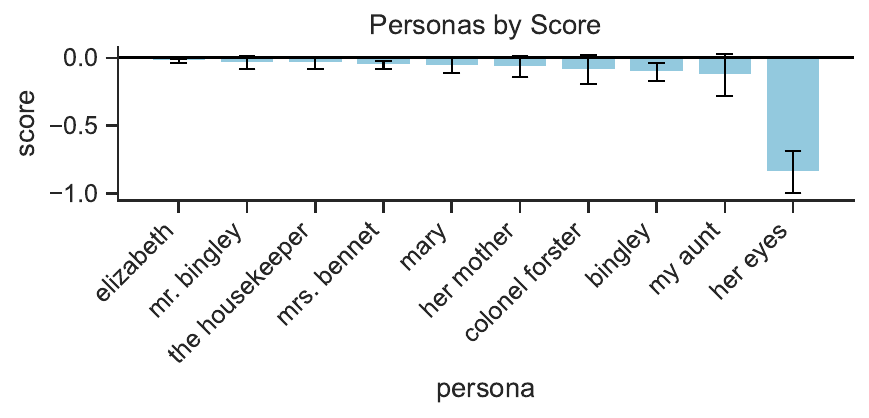}
    \caption{Means and standard deviations of power scores across 20 bootstrapped samples of \textit{Pride and Prejudice}, using \packageName's sampling and visualization functions. Results are shown for three custom pronoun groups captured using \packageName's entity discovery and coreference resolution capabilities.}
    \label{figure:pride-and-prejudice-barplot}
\end{figure}

\begin{figure}[t]
    \centering
    \includegraphics[width=2.24cm]{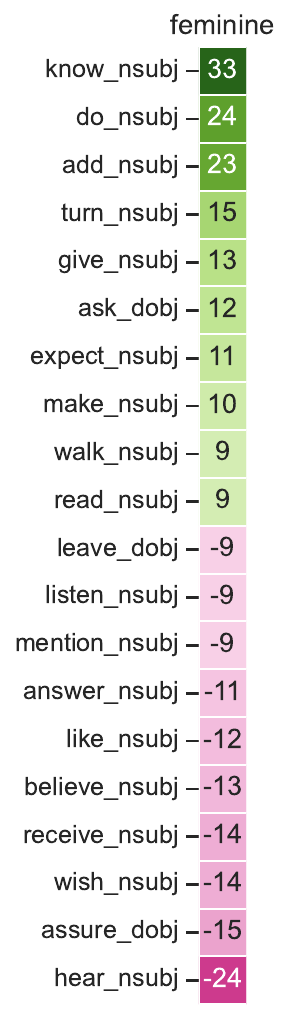}
    \includegraphics[width=2.545cm]{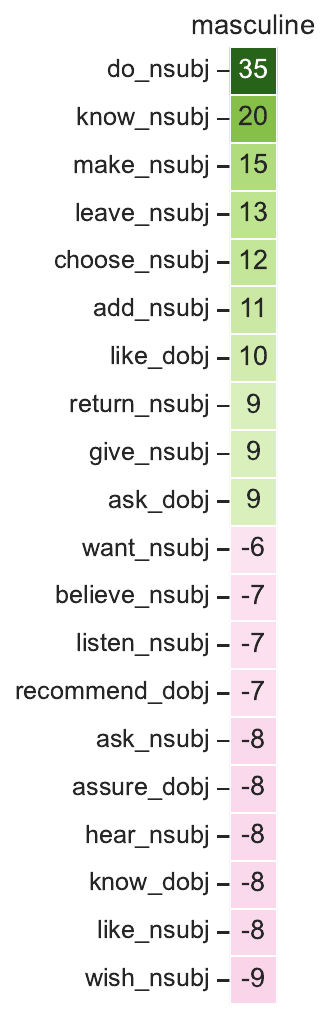}
    \includegraphics[width=2.55cm]{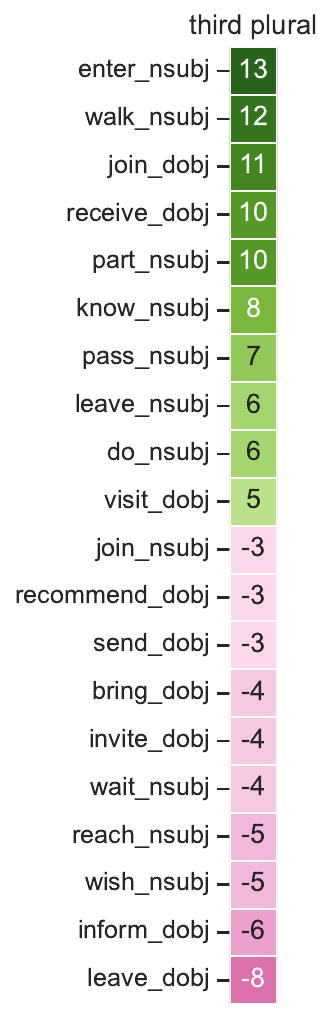}
    \caption{Verb counts for pronoun groups in \textit{Pride and Prejudice}, using \packageName's visualization functions. Green cells indicate verbs where the pronoun group \textbf{has} power, while pink cells indicate verbs where the pronoun group \textbf{lacks} power. Labels include both the verb and the position (\textit{nsubj} or \textit{dobj}) of the pronoun group, and the pronoun groups were captured via \packageName's customizable entity discovery function.}
    \label{figure:pride-and-prejudice-top-words}
\end{figure}

\subsection{Gender Differences in \textit{Pride and Prejudice}}

Jane Austen's 1813 novel \textit{Pride and Prejudice} is famous for its depiction of gender and class relations in 19th century England.
Using the entity recognition and coreference resolution capabilties of \packageName, we can identify which characters are framed as having or lacking power, and by examining the power relations between classes of third person pronouns, we can trace how gender hierarchies are enacted and subverted by Austen, through the actions of her characters.

\paragraph{Results}

Figure \ref{figure:pride-and-prejudice-barplot} shows the entities identified by \packageName that have the highest and lowest power scores, using the lexicon from \citet{sap2017connotation}.
Characters like Miss Lucas have higher power scores, while characters like Elizabeth have lower power scores.
By using \packageName's functionality to pull out the documents contributing to an entity's score, we find that a mistaken entity, ``her eyes,'' indeed often occurs in low power roles, as in ``Miss Bingley fixed her eyes on face,'' providing an intuitive validation of the results.
This plot also shows the standard deviation across bootstrapped samples of the novel, indicating the overlapping instability of many of the power scores for this single novel.

Figure \ref{figure:pride-and-prejudice-top-words} shows the lexicon verbs contributing most frequently to each pronoun group's power score. 
For example, we see that feminine pronouns are frequently used as subjects of the verb ``hear''---emphasizing women's low-power role of waiting to hear news.
We also observe that while feminine pronouns are often placed in high-power positions at rates similar to masculine pronouns, they have higher frequencies for low-power positions. 
In other words, in Austen's world, masculine and feminine entities both engage in high-power actions, but feminine entities engage in more low-power actions.
Arguably, though, some of the low-power positions are used by the feminine entities to obtain power, e.g., by ``hearing'' news or eavesdropping on others, the feminine entities can learn information that informs their future decisions and strategies.

\section{Ethics and Broader Impact}

\packageName comes with some risks and limitations.
This package is targeted only at English-language texts; we have not included non-English lexica in the tool nor do we expect the parsing, named entity recognition, and coreference resolution to directly translate to other languages.
While related lexica exist for non-English languages (e.g., \citet{klenner-etal-2017-stance} (German), \citet{rashkin-etal-2017-multilingual} (extension to 10 European languages)), the generalizability of \packageName is limited to English-language settings.

The results of \packageName are only as reliable as the corpora and lexica used as input (and their relationships to one another).
We have emphasized interpretability in designing this package, encouraging users to examine their results at different levels of granularity.
However, there are still dangers of biases ``baked-in'' to the data, via its sampling and curation, or to the lexica, in the choice of terms and their annotations by human workers.
Lexica that are useful in one setting are not always useful in other settings, and we encourage researchers to validate their lexica on their target corpora.

Drawing from a framework describing the roles for computational tools in social change \citep{abebe2020roles}, we believe that \packageName can fill multiple important roles.
First, it can act as a \textit{diagnostic}, measuring social biases and hierarchies across large corpora, as in \citet{mendelsohn2020framework} where dehumanization of LGBTQ+ people was measured across news datasets and time. 
\packageName can also act as a \textit{formalizer}, allowing researchers to examine the specific words used by authors, adding concrete and fine-grained evidence to the constructions of broader patterns, as in \citet{antoniak2019narrative} where the supporting words were used to characterize healthcare roles during childbirth.
Finally, \packageName can act as a \textit{synecdoche} by bringing new attention and perspectives to long-recognized problems, as in \citet{sap2017connotation} where renewed attention was given to gender biases in film.

\section{Acknowledgements}

We thank Hannah Rashkin for the invaluable feedback on the paper and for creating the original connotation frames.
We also thank our anonymous reviewers whose comments helped us improve both this paper and \packageName.

\bibliography{ref}
\bibliographystyle{acl}

\clearpage
\appendix

\section{Appendix: GPT-3.5 Generation Setup}
\label{sec:appendix-gpt3}
Here we further detail our steps to evaluate our framework as discussed in Section~\ref{subsec:power-dyads}. 
To generate characters with clear roles, names, and power differences, we used 32 dyadic relation pairs with explicit power asymmetry in our prompt.
The full list of relations are shown in Table~\ref{tab:dyadic_roles}.
The following is an example of the prompt used to generate such stories: 
\begin{quote}
    Tell me a short story about a doctor and a patient, and give them names.
    
    doctor's name:
\end{quote}
Using text-davinci-003 model, we generated 3 stories per pair with temperature set to 0.7 and max tokens set to 256. 
After cleaning ill-formatted results, we analyzed a total of 85 stories. 
A few examples of the generated stories along with the power scores of the characters are shown in Table~\ref{tab:gpt3_stories}.

\begin{table}[h]
    \centering
    \begin{tabular}{ll}
        (doctor, patient) & (teacher, student)\\
        (interviewer, interviewee) & (parent, child)\\
        (employer, employee) & (boss, subordinate)\\
        (manager, worker) & (landlord, tenant)\\
        (judge, defendant) & (supervisor, intern)\\
        (therapist, client) & (owner, customer)\\
        (mentor, mentee) & (politician, voter)\\
        (rich, poor) & (elder, younger)\\
        (artist, critic) & (host, guest)\\
        (preacher, parishioner) & (expert, novice)\\
        (counselor, advisee) & (coach, athlete)\\
        (lender, borrower) & (king, subject)\\
        (seller, buyer) & (umpire, player)\\
        (printer, writer) & (sponsor, sponsored)\\
        (producer, actor) & (legislator, citizen)\\
        (commander, soldier) & (executor, beneficiary)
    \end{tabular}
    \caption{Full list of dyadic roles}
    \label{tab:dyadic_roles}
\end{table}

\begin{table*}[t]
    \centering
    \begin{tabularx}{\linewidth}{XXl}
    \toprule
        \multicolumn{3}{X}{GPT3.5 Generated Story} \\
        Dyadic Roles & Power Scores & Power Score Difference\\
        \midrule
        \multicolumn{3}{p{14cm}}{Alan was a successful CEO of a large company. He had a sharp eye for detail and was known for his strict yet fair leadership style. One day, Alan noticed that one of his subordinates, Zara, was not performing up to the standards he expected. He called her into his office and asked her why her work wasn't up to par. Zara admitted that she was feeling overwhelmed with her workload and was having trouble keeping up. She asked Alan if he could help her prioritize her tasks and provide her with additional resources to help her complete her work. Alan was impressed by Zara's initiative and agreed to help her. He rearranged her tasks and assigned her additional help from other staff members. He also gave her some helpful tips on how to manage her workload more effectively. Zara was grateful for Alan's help and was able to complete her work on time and to a high standard. Alan was pleased with her performance and decided to reward her with a promotion.  From then on, Zara and Alan had an excellent working relationship. Zara's hard work and resilience was an example to her colleagues, and Alan's willingness to help her out was a}\\
        (boss, subordinate) & Alan: 0.4, Zara: -0.07 & 0.47\\
        \hline
        \multicolumn{3}{p{14cm}}{Mr. Jones was a landlord who owned an apartment building in a small town. He had been a landlord for many years, and he was known for his strict rules and no-nonsense attitude. Ms. Lopez was a single mother who had recently moved into the building. She was looking for a safe place to raise her young daughter, and Mr. Jones' building seemed to fit the bill. She was a bit intimidated by his strict rules, but she was willing to abide by them if it meant living in a safe and secure place. At first, things went smoothly between Mr. Jones and Ms. Lopez. He made sure the building was well maintained, and she was a model tenant, paying her rent on time every month. However, one day Ms. Lopez noticed that her hot water heater had stopped working. She immediately contacted Mr. Jones to let him know, but he refused to repair it. She was frustrated and angry, but she didn't want to risk getting evicted, so she kept quiet. After a few weeks, the hot water heater finally broke down completely and Ms. Lopez had no choice but to contact Mr. Jones again.}\\
        (landlord, tenant)  & Mr. Jones: 0.42, Ms. Lopez: 0.09 & 0.33\\
        \hline
        \multicolumn{3}{p{14cm}}{Paul was an experienced interviewer, but Emily was his most challenging interviewee yet. He had seen her resume online and was impressed with her qualifications. He had high hopes for her. When Emily arrived for the interview, Paul immediately noticed her confidence. She was wearing a suit and had a bright smile on her face. She was clearly prepared and ready for the interview.  Paul asked Emily several questions about her past experience and qualifications. She answered each one with poise and detail. She was articulate and could back up her answers with examples.  Paul was very impressed with Emily's performance. He could tell she was a strong candidate and he was confident she would do well in the job.  At the end of the interview, Paul thanked Emily for her time and wished her luck. He was sure she would be a great addition to the company.  Emily left the interview feeling pleased with how it went. She was confident she had done her best and was sure she had impressed Paul.  The next day, Paul gave Emily an offer for the job. She happily accepted and started a few weeks later.  Paul and Emily had a successful working}\\
        (interviewer, interviewee) & Paul: 0.07, Emily: 0.25 & -0.17\\
    \bottomrule
    \end{tabularx}
    \caption{Examples of GPT3.5 generated stories and the power scores of the characters}
    \label{tab:gpt3_stories}
\end{table*}

\end{document}